\documentclass[twoside]{IEEEtran}
\usepackage{amsmath,amsthm,amssymb,amsfonts,amscd,graphicx}
\usepackage{graphics}

\theoremstyle{definition}
\newtheorem{definition}{Definition}
\newtheorem{algorithm}{Algorithm}
\newtheorem{example}{Example}

\newcommand{\R}{\mathbb{R}}

\renewcommand{\d}{\partial}

\begin{document}

\title{A Multivariate Hawkes Process with Gaps in Observations}
\author{Triet M. Le\thanks{NGA Research, National Geospatial-Intelligence Agency, 7500 GEOINT Dr., Springfield, VA 22150, Email: Triet.M.Le@nga.mil. 
}}
\maketitle

\begin{abstract}
Given a collection of entities (or nodes) in a network and our intermittent observations of activities from each entity, an important problem is to learn the hidden edges depicting directional relationships among these entities. Here, we study causal relationships (excitations) that are realized by a multivariate Hawkes process. The multivariate Hawkes process (MHP) and its variations (spatio-temporal point processes) have been used to study contagion in earthquakes, crimes, neural spiking activities, the stock and foreign exchange markets, etc.  In this paper, we consider the multivariate Hawkes process with gaps in observations (MHPG). We propose a variational problem for detecting sparsely hidden relationships with a multivariate Hawkes process that takes into account the gaps from each entity. We bypass the problem of dealing with a {\em large} amount of missing events by introducing a {\em small} number of unknown boundary conditions. In the case where our observations are sparse (e.g. from $10\%$ to $30\%$), we show through numerical simulations that robust recovery with MHPG is still possible even if the lengths of the observed intervals are small but they are chosen accordingly. {\em The numerical results also show that the knowledge of gaps and imposing the right boundary conditions are very crucial in discovering the underlying patterns and hidden relationships.}
\end{abstract}


\section{Introduction}
Point processes have demonstrated to be promising tools for extracting dynamic patterns and discovering hidden relationships in event data. Variations of (spatio-temporal, univariate, multivariate) point processes  have been applied to event data from many different fields in science. For instance, self-exciting point processes have been used in seismology to model contagion of earthquakes \cite{ogata1984transition, ogata1988statistical, ogata1999seismicity, zhuang2002stochastic}, and in anthropology to study the spread of crimes and violence acts \cite{mohler2011self, lewis2012self, sidebottom2012repeat, short2014gang}. The multivariate Hawkes process (a parametric version of the self-exciting point process) has been applied to financial data to study contagion and influential entities in financial networks \cite{ait2010modeling, azizpour2008self, bowsher2007modeling, embrechts2011multivariate, embrechts2016hawkes}, and also in social media networks \cite{stomakhin2011reconstruction, zipkin14point, hall2014tracking, masuda2013self, etesami2016learning}. The multivariate Hawkes process with inhibition has also been used in neuroscience to make inference of functional connectivity from neural spiking activities \cite{kim2011granger, reynaud2013inference}, among others. The common assumption in these work is that all events over a long-enough time interval of interest are observed. However, for reasons associated to the environment, etc., events are only observed intermittently. Thus the challenges are: 1) how to recover robustly the underlying parameters in the presence of gaps; 2) how to distribute the gaps for optimal recovery given the available resources.

In technical terms, let $N$ be the number of entities (or nodes) within a network. For each entity $m$ ranging from $1$ to $N$, let $E_m = \{t_{m,i}\}$ be the set of events that entity $m$ generates in some interval of interest, say $(0,T]$. Here each $t_{m,i}$ in $(0,T]$ (assuming $t_{m,i-1} < t_{m,i}$) represents a time-stamp when an event from entity $m$ occurs. We recall the following definitions of point processes.

\begin{definition}[Poisson Process \cite{daley2003introduction}]
For each entity $m$, denote the finite collection of disjoint intervals in $(0,T]$ by $\{(c_{m,k},d_{m,k}]\}_{k=1}^{K_m}$. Let $N(c_{m,k},d_{m,k}]$ be the number of observed events from entity $m$ that are contained in $(c_{m,k},d_{m,k}]$. We say the collection of observed events in $E_m$ follows a {\bf homogeneous} Poisson process with some constant intensity $\lambda_m\ge 0$ if $N(c_{m,k},d_{m,k}]$ follows a Poisson distribution with mean $\lambda_m(d_{m,k}-c_{m,k})$. In other words, the following probability holds
\begin{equation}\label{hpp_def_eq1}
\begin{split}
P(N(c_{m,k},d_{m,k}] &= n) \\
&= \frac{[\lambda_m(d_{m,k}-c_{m,k})]^{n} e^{-\lambda_m(d_{m,k}-c_{m,k})}}{n!}.
\end{split}
\end{equation}
Suppose now instead of a constant $\lambda_m$, we have a positive (integrable) function $\lambda_m:(0,T]\rightarrow \R^+$.  Then we say the set of events $E_m$ follows an {\bf inhomogeneous} Poisson process with intensity function $\lambda_m(t)$ if
\begin{equation}\label{ipp_def_eq1}
\begin{split}
P(N(c_{m,k},d_{m,k}] = n) 
= \frac{[\Lambda_{m,k}]^{n} e^{-\Lambda_{m,k}}}{n!},
\end{split}
\end{equation}
where
$$
\Lambda_{m,k} = \int_{c_{m,k}}^{d_{m,k}} \lambda_m(t)\ dt.
$$
\end{definition} 
The interpretations of \eqref{hpp_def_eq1} (or \eqref{ipp_def_eq1}) are as follows:
\begin{enumerate}
\item The number of events in $(c_{m,k},d_{m,k}]$ follows a Poisson distribution with mean and variance $\lambda_m(d_{m,k}-c_{m,k})$ (or $\int_{c_{m,k}}^{d_{m,k}}\lambda_m(t)\ dt$).
\item The number of events in disjoint intervals or from different entities are independent random variables. In other words, an occurrence of an event from entity $m$ has no influence on future events from itself or from any other entities.
\end{enumerate} 

The multivariate Hawkes process introduces {\em directional dependencies among entities and events} into the definition of the (conditional) intensity function $\lambda_m(t)$. The word {\em \lq conditional\rq} is used because $\lambda_m(t)$ is conditioned on prior events. 

\begin{definition}[Multivariate Hawkes Process \cite{hawkes1971spectra}]
We say the collection of observed events in $E_m, m=1,\cdots, N,$ follows a multivariate Hawkes process if for all $t$ in  $(0,T]$, the conditional intensity function (CIF) $\lambda_m(t)$ for entity $m$ is given by
\begin{equation}\label{cif_eq}
\lambda_m(t) = u_m + \sum_{n=1}^N a_{m,n} \sum_{\substack{t_{n,j} \in E_n\\ t_{n,j}<t}} b_me^{-b_m(t-t_{n,j})}.
\end{equation}
\end{definition}
Here, the background $u_m\ge 0$ is a homogeneous Poisson process, and it is included here to promote independent random events. Since $\int_0^\infty b_me^{-b_mt}\ dt=1$ for $b_m>0$, the matrix $a=(a_{m,n})_{N\times N}$ with the entry $a_{m,n}\ge 0$ depicts how an event $t_{n,j}$ from entity $n$ will trigger or excite future events from entity $m$. A multivariate Hawkes process is stationary if and only if the largest eigenvalue of the matrix $a$ in absolute value is strictly bounded above by $1$ \cite{daley2003introduction}. $1/b_m$ (the width of the exponential function) is the timescale providing the likelihood when the next event from entity $m$ occurs.

To incorporate inhibition into the multivariate Hawkes process, one allows $a_{m,n}$ to be negative and the conditional intensity $\lambda_m(t) \ge 0$ can be defined as  \cite{reynaud2013inference} 
$$
\lambda_m(t) = G\left(u_m + \sum_{n=1}^N a_{m,n} \sum_{\substack{t_{n,j} \in E_n\\ t_{n,j}<t}} b_me^{-b_m(t-t_{n,j})} \right),
$$
where for example $G(x) = \max(0,x)$ or $G(x) = e^x$. In this paper we focus on the case where $a_{m,n}$ is non-negative.

It is possible to consider an inhomogeneous Poisson process for the background, and to have a different timescale or mode of excitation for each pair of entities. For simplicity, we focus on the single mode of excitation case given by \eqref{cif_eq}. Also, each event $t_{m,i}$ may have a different mark or jump size $M_{m,i}\in [0,1]$. In other words, $\lambda_m(t)$ can be defined as
$$
\lambda_m(t) = u_m + \sum_{n=1}^N a_{m,n} \sum_{\substack{t_{n,j} \in E_n\\ t_{n,j}<t}} M_{n,j}b_me^{-b_m(t-t_{n,j})}.
$$
Here we consider all events to be the same, namely $M_{m,i} = 1$.

\begin{figure}
\centering
$\underset{(a)}{\includegraphics[scale=0.275]{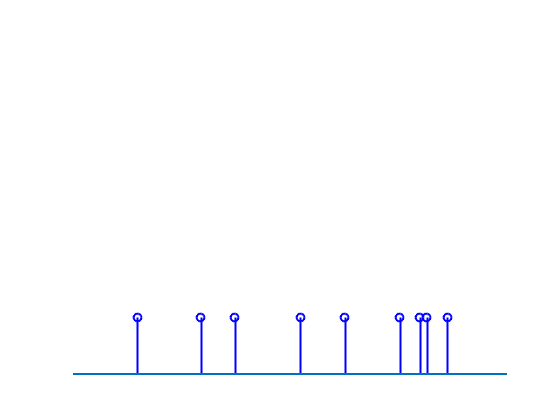}}$
$\underset{(b)}{\includegraphics[scale=0.275]{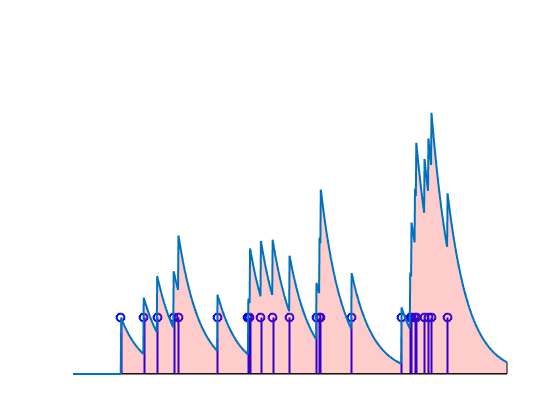}}$
\caption{Simulations of two univariate point processes in the interval $(0,T]$, $T=10$: (a): Homogeneous Poisson process with constant intensity $\lambda = 1$, (b): A univariate Hawkes process depicting the dynamic of $\lambda(t)$ given in \eqref{uvh_eq} as events (blue spikes) evolve. Here $u=1, a=0.5$ and $b=2$.}
\label{fig1}
\end{figure}

Figure \ref{fig1} shows simulations of two univariate point processes ($N=1$) in the interval $(0,T]$, with $T=10$. Figure \ref{fig1}(a) shows a homogeneous Poisson process with constant $\lambda=1$, and Figure \ref{fig1}(b) shows a (univariate) Hawkes process with $u=1, a=0.5$ and $b=2$. Recall the CIF of a univariate Hawkes process (Equation \eqref{cif_eq} with $N=1$) is defined as
\begin{equation}\label{uvh_eq}
\lambda(t) = u + a \sum_{0<t_i<t}be^{-b(t-t_i)}.
\end{equation}
Note the dynamics of $\lambda(t)$ in Figure \ref{fig1}(b) as events (in blue spikes) evolve. Based on the definition of a homogeneous Poisson process, we expect that events are uniformly distributed (as evident in Figure \ref{fig1}(a)). The same phenomenon doesn't hold for a Hawkes process whenever $ab>0$. This is evident in Figure \ref{fig1}(b) as events are more clustered as a result of self-excitation and hence there are more burstiness in the intensity function.

\begin{figure}
\centering
\includegraphics[scale=0.425]{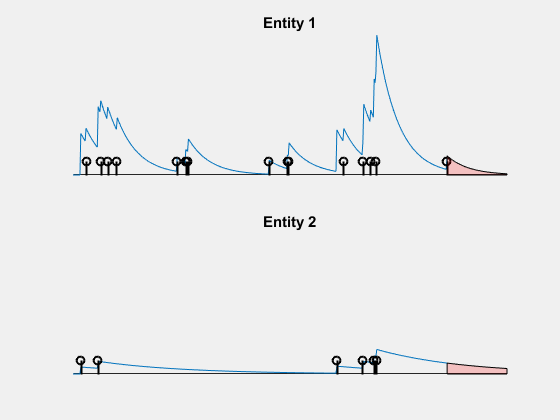}
\caption{A simulation of a multivariate Hawkes process ($N=2$): In this Figure, entity $2$ is very influential toward entity $1$ (since $a_{1,2} = 0.75$) and entity $1$ has no influence on entity $2$ (since $a_{2,1} = 0$). Both entities have the same amount of self-excitation ($a_{1,1} = a_{2,2} = 0.25$).}
\label{fig2_1}
\end{figure}

Figure \ref{fig2_1}  shows a simulation of a multivariate Hawkes process ($N=2$) with $u = \begin{pmatrix} 0.1 \\ 0.1\end{pmatrix}, a = \begin{pmatrix}0.25 & 0.75 \\ 0 & 0.25 \end{pmatrix},  b = \begin{pmatrix}10 \\ 1\end{pmatrix}$. $a_{1,2} = 0.75$ implies that events from entity $2$ is very contagious toward entity $1$. On the other hand, events from entity $1$ has no influence on entity $2$ since $a_{2,1} = 0$. These effects can be seen in the evolution of the CIFs (in blue). An event from entity $2$ creates a jump in the CIF of entity $1$ which causes a series of events to follow, but not vice versa.

\begin{figure}
\begin{center}
\includegraphics[scale=0.425]{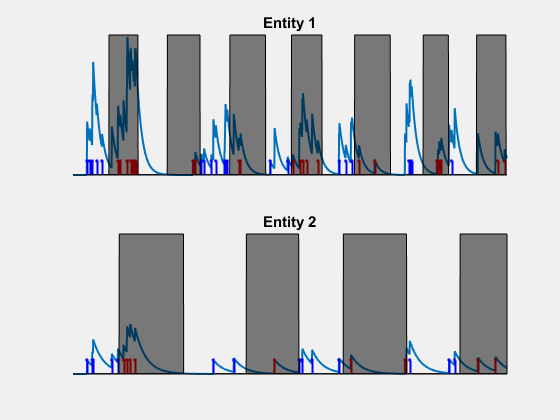}
\caption{A simulation (using the same parameters as in Figure \ref{fig2_1}) showing the intermittent observations for each entity. Blue spikes are observed events and red spikes are unobserved events.}
\label{fig3}
\end{center}
\end{figure}

In this paper, we consider the case where one has intermittent observations (and hence gaps). Figure \ref{fig3} shows an example of intermittent observations for a network of two entities. The shaded intervals represent the observational gaps. We observe events in blue and do not observe events in red from each entity. For this simulation, we use the same parameters as in Figure \ref{fig2_1}. 

 For each entity $m$, let $\{(c_{m,k},d_{m,k}]\}_{k=1}^{K_m}$ be the collection of disjoint observed intervals (e.g. unshaded intervals in Figure \ref{fig3}) that are contained in $(0,T]$, and let $O_m = \{t_{m,i}\}$ be the corresponding partially observed events (e.g. events in blue from Figure \ref{fig3}). For each $t$ that belongs to one of the observed intervals $(c_{m,k},d_{m,k}]$ we replace the original CIF for entity $m$  in \eqref{cif_eq} with

\begin{equation}\label{cif_approx}
\begin{split}
\bar{\lambda}_m(t) &= u_m + (\bar{\lambda}_m(c_{m,k}) - u_m)e^{-b_m(t - c_{m,k})} \\
&+  \sum_{n=1}^N a_{m,n}\sum_{\substack{t_{n,j}\in O_n\\ c_{m,k} < t_{n,j}<t}}b_m e^{-b_m(t-t_{n,j})},
\end{split}
\end{equation}
where $\bar{\lambda}_{m,k} := \bar{\lambda}_m(c_{m,k})$ are extra unknowns. By an abuse of notation, denote the vector $\bar{\lambda} := \{\lambda_{m,k}\}$. To learn the underlying parameters $u,a$ and $b$, we propose to solve the following minimization problem,
\begin{equation}\label{J_orig}
\begin{split}
J(u,a,b,\bar{\lambda}) &= \sum_{m=1}^N \sum_{k=1}^{K_m}\left[\int_{c_{m,k}}^{d_{m,k}}\bar{\lambda}_m(t)\ dt \right.\\
&\left.- \sum_{t_{m,i}\in (c_{m,k},d_{m,k}]}\log\left(\bar{\lambda}_{m}(t_{m,i})\right)\right]\\
&+ G(a) + H(\bar{\lambda}),
\end{split}
\end{equation}
where $G(a)$ and $H(\bar{\lambda})$ are appropriate constraints or regularizations on the matrix $a$ and the vector $\bar{\lambda}$.

The paper is organized as follows. In Section \ref{partial} we go over a formulation to incorporate the gaps into the modeling of \eqref{cif_approx} and go over the variational model \eqref{J_orig} with appropriate regularizations and constraints to learn the underlying parameters $u, a$ and $b$. In Section \ref{num_result}, we provide a numerical study using simulated data to show that the proposed method robustly recovers the underlying parameters in the presence of large amount of missing events ($\ge 70\%$). A detailed description of the numerical implementation for computing a minimizer of \eqref{J_orig} (see Algorithm \eqref{param_est_alg}) is outlined in the appendix.

To the author's knowledge, there isn't any work in the literature that addresses intermittent observations for point processes in a continuous setting.


\section{Multivariate Hawkes Process with Gaps (MHPG)}\label{partial}

For each $m=1,\cdots, N$, let $E_m = \{t_{m,i}\}$ be the complete set of events from entity $m$ that are contained in $(0,T]$. Recall from \eqref{cif_eq} that the CIF for entity $m$ is given by
\begin{equation}\label{cif_eq2}
\lambda_m(t) = u_m + \sum_{n=1}^N a_{m,n} \sum_{\substack{t_{n,j} \in E_n\\ t_{n,j}<t}} b_me^{-b_m(t-t_{n,j})}.
\end{equation}
It can be shown that $\lambda_m(t)$ satisfies the following mean-reverting dynamics
\begin{equation} \label{ode_cif_eq}
d\lambda_m(t) = b_m(u_m-\lambda_m(t))dt  + \sum_{n=1}^N a_{m,n} b_m dN_n(t),
\end{equation} 
for $t\in (0,T]$, with the boundary condition $\lambda_m(0) = u_m$. In general, the solution to \eqref{ode_cif_eq} has the form
\begin{equation*}
\begin{split}
\lambda_m(t) &= u_m + (\lambda_m(0) - u_m) e^{-b_mt} \\
&+ \sum_{n=1}^N a_{m,n} \sum_{\substack{t_{n,j} \in E_n\\ t_{n,j}<t}} b_me^{-b_m(t-t_{n,j})}.
\end{split}
\end{equation*}

{\bf MHP}: Given $E_m$, for $m=1,\cdots, N$, the task is to learn the parameters $u = (u_m)_{N\times 1}$, $a = (a_{m,n})_{N\times N}$ and $b = (b_m)_{N\times 1}$. The common approach is to minimize the $(-)$log-likelihood functional \cite{daley2003introduction}:
\begin{equation}\label{J_mhp}
\begin{split}
\min_{u,a,b} &\left\{L(u,a,b) =\sum_{m=1}^N\left[ \int_0^T \lambda_m(t)\ dt \right.\right.\\
&\left.\left.- \sum_{t_{m,i}\in E_m}\log(\lambda_m(t_{m,i}))\right]  + G(a)\right\},
\end{split}
\end{equation}
with the constraint $u_m\ge 0$ and $b_m\ge 0$. The second term $G(a)$ is the prior or regularization on the matrix $a$. For instance, to impose sparsity on interactions, one can use the LASSO constraint \cite{tibshirani1996regression} $G(a) =\mu \sum_{m,n} |a_{m,n}|$ for some $\mu>0$ .

Suppose now that we do not have complete observations, that is let $\{(c_{m,k},d_{m,k}]\}_{k=1}^{K_m}$ be the collection of disjoint observed intervals for entity $m$ that are contained in $(0,T]$. Here we assume that $t_{n,j}\neq c_{m,k}$ for all events $t_{n,j}$ and boundary values $c_{m,k}$. Let $O_m$ be the set of the corresponding partially observed events, that is
$$
O_m = E_m \cap \left[\cup_{k=1}^{K_m} (c_{m,k},d_{m,k}] \right].
$$

Let $t\in (c_{m,k},d_{m,k}]$ and recall from \eqref{cif_eq2}, we have (assuming $\lambda_m(0) = u_m$)

\begin{eqnarray}
\label{cif_eq_orig}
\lambda_m(t) &=& u_m +  \sum_{n=1}^N a_{m,n}\sum_{\substack{t_{n,j}\in E_n\\ t_{n,j}<t}}b_m e^{-b_m(t-t_{n,j})}\\
&=& u_m +  \sum_{n=1}^N a_{m,n}\sum_{\substack{t_{n,j}\in E_n\\ t_{n,j}<c_{m,k}}}b_m e^{-b_m(t-t_{n,j})} \\
&+& \sum_{n=1}^N a_{m,n}\sum_{\substack{t_{n,j}\in E_n\\ c_{m,k} < t_{n,j}<t}}b_m e^{-b_m(t-t_{n,j})}.
\label{lambda_eq0}
\end{eqnarray}
Since by  \eqref{cif_eq}, 
$$
\lambda_m(c_{m,k}) = u_m + \sum_{n=1}^N a_{m,n}\sum_{\substack{t_{n,j}\in E_n\\ t_{n,j}<c_{m,k}}}b_m e^{-b_m(c_{m,k}-t_{n,j})}.
$$
Substituting $\lambda_m(c_{m,k})$ into \eqref{lambda_eq0}, we get
\begin{equation}\label{pcif_eq1}
\begin{split}
\lambda_m(t) &= u_m + (\lambda_m(c_{m,k}) - u_m)e^{-b_m(t - c_{m,k})} \\
&+  \sum_{n=1}^N a_{m,n}\sum_{\substack{t_{n,j}\in E_n\\ c_{m,k} < t_{n,j}<t}}b_m e^{-b_m(t-t_{n,j})}.
\end{split}
\end{equation}
The boundary value $\lambda_{m,k} := \lambda_m(c_{m,k})$ is an extra unknown since it may depend on the unobserved events that are contained in the gap $(d_{m,k-1},c_{m,k}]$. The third term on the right-hand side of the last equation is summing over (observed and unobserved) events from entity $n$ that are contained in $(c_{m,k},d_{m,k}]$. Clearly, if $m$ and $n$ have the same observed intervals, then we have 
\begin{equation}\label{iden_sets_eq}
\begin{split}
&\{t_{n,j} \in E_n: c_{m,k} < t_{n,j}\le d_{m,k}\} \\
&= \{t_{n,j}\in O_n:  c_{m,k} < t_{n,j}\le d_{m,k}\}.
\end{split}
\end{equation}
In general, we consider the following approximation of the CIF for entity $m$:
\begin{equation}\label{pcif_eq}
\begin{split}
\bar{\lambda}_m(t) &= u_m + (\lambda_m(c_{m,k}) - u_m)e^{-b_m(t - c_{m,k})} \\
&+  \sum_{n=1}^N a_{m,n}\sum_{\substack{t_{n,j}\in O_n\\ c_{m,k} < t_{n,j}<t}}b_m e^{-b_m(t-t_{n,j})}.
\end{split}
\end{equation}
Note that the representation of $\bar{\lambda}_m(t)$ in \eqref{pcif_eq} is exactly equal to $\lambda_m(t)$ in \eqref{pcif_eq1} whenever \eqref{iden_sets_eq} holds. For the general case where  the observed intervals for $m$ and $n$ are not identical, then \eqref{pcif_eq} provides an approximation to $\lambda_m(t)$. In Section \ref{num_result}, we show that by taking the intersection of the observed intervals and remove events that do not belong to this intersection, one can achieve better reconstructions.

{\bf MHPG}: Denote $\bar{\lambda}_{m,k} := \lambda_m(c_{m,k})$ and by an abuse of notation denote $\bar{\lambda} := \{\bar{\lambda}_{m,k}\}$. We propose to learn the parameters $u,a, b$ and $\bar{\lambda}$ by minimizing the following functional:
\begin{equation}\label{J_mhpg}
\begin{split}
J(u,a,b,\bar{\lambda}) &= \sum_{m=1}^N \sum_{k=1}^{K_m}\left[\int_{c_{m,k}}^{d_{m,k}}\bar{\lambda}_m(t)\ dt\right.\\
 &- \left.\sum_{t_{m,i}\in (c_{m,k},d_{m,k}]}\log\left(\bar{\lambda}_{m}(t_{m,i})\right)\right]\\
&+ G(a) + H(\bar{\lambda}),
\end{split}
\end{equation}
where $G(a)$ and $H(\bar{\lambda})$ are priors (or regularizations) on $a$ and $\bar{\lambda}$ respectively. If all entities have the same observed intervals, then using techniques from \cite{daley2003introduction}, one can show that $J$ from \eqref{J_mhpg} is the $(-)$log-likelihood function. From the graph/network point of view, the LASSO constraint \cite{tibshirani1996regression} on the matrix $a$,
\begin{equation}\label{G_eq}
G(a) = \mu\sum_{m,n=1}^N |a_{m,n}|,
\end{equation}
enforces sparsity on $a$. In other words, each entity only interacts with a few other entities within the network.

Theorem 4.1 from \cite{dassios2011dynamic} provides a theoretical result for the marginal distribution of $\{\lambda_m(t)\}_{t>0}$. The (-)log of this distribution can be used to define $H(\bar{\lambda})$. For instance, for a univariate Hawkes process \eqref{uvh_eq}, $\{\lambda(t)\}_{t>0}$ follows a shifted Gamma distribution
\begin{equation}\label{uvh_cif_dis}
\{\lambda(t)\}_{t>0}\approx u + \mbox{Gamma}\left(\frac{u}{b}, \frac{1-a}{ab} \right).
\end{equation}
This implies 
\begin{equation}\label{uvh_mean_var_eq}
\begin{split}
\mbox{mean}\left(\{\lambda(t)\}_{t>0}\right) &= u + \frac{ua}{1-a}, \mbox{ and }\\
\mbox{var}\left(\{\lambda(t)\}_{t>0}\right) &= \frac{ua^2b}{(1-a)^2}.
\end{split}
\end{equation}
Clearly as $a\rightarrow 1$ both the mean and variance converge to $\infty$. Note that a small change in the parameter $a$ produces a large change in both the mean and variance. For a multivariate Hawkes process the distribution of $\{\lambda_m(t)\}_{t>0}$  has a much more complicated form. Thus, we consider instead the following constraint on $\bar{\lambda}_{m,k}$:
\begin{equation}\label{lambda_cons}
u_m \le \bar{\lambda}_{m,k} \le Cu_m,
\end{equation}
for some $C\ge 1$. This can be viewed as having $\bar{\lambda}_{m,k}$ following a uniform distribution on $[u_m,Cu_m]$.


The functional $J$ in \eqref{J_mhpg} is not convex, in particular, with respect to $b$. There are numerous successful numerical schemes that have been proposed to compute a minimizer for non-convex functionals. See for instance PALM \cite{bolte2014proximal}, or Block Prox-Linear Method \cite{xu2014globally}, among others. The method we use here follows PALM but instead of using gradient descend which is very slow in practice we use the fixed point method. Below is a summary of the proposed algorithm where the detail of the numerical implementation is given in the appendix

\begin{algorithm}[Algorithm for parameter estimation]\label{param_est_alg}
Given 
$$
O_m = \{t_{m,i}\}\subset \cup_{k=1}^{K_m}(c_{m,k},d_{m,k}], m=1,\cdots, N, 
$$
some $\mu>0$ and $dt=$ small.
\begin{enumerate}
\item Initial guess: $u^0_m = 1, a^0_{m,n} = 0.5/N, b_m = 1000, \bar{\lambda}^0 = \{\bar{\lambda}^0_{m,k} = u^0_m\}.$
\item Suppose $u^\ell, a^\ell, b^\ell$ and $\bar{\lambda}^\ell$ are known.
\item While not convergent
\begin{enumerate}
\item Compute $u^{\ell +1}$ using $u^\ell, a^\ell, b^\ell$ and $\bar{\lambda}^\ell$ via \eqref{u_eq}.
\item Compute $a^{\ell+1}$ using $u^{\ell+1}, a^\ell, b^\ell$ and $\bar{\lambda}^\ell$ via \eqref{a_eq}.
\item Compute $b^{\ell+1}$ using $u^{\ell+1}, a^{\ell+1}, b^\ell$ and $\bar{\lambda}^\ell$ via \eqref{b_eq}.
\item Compute $\bar{\lambda}^{\ell+1}$ using $u^{\ell+1}, a^{\ell+1}, b^{\ell+1}$ and $\bar{\lambda}^\ell$ via \eqref{lambda_eq} using the constraint \eqref{lambda_cons}.
\end{enumerate}
\item End while. 
\end{enumerate}
\end{algorithm}


Although Equation \eqref{pcif_eq1} can be derived directly from Equation \eqref{ode_cif_eq} for $t\in (c_{m,k},d_{m,k}]$, the technique described in Equations \eqref{cif_eq_orig}-\eqref{lambda_eq0} can be applied to a much more general case. Transforming Equation \eqref{cif_eq_orig} to Equation \eqref{pcif_eq} is possible because of the fact that $e^{-bt}$ satisfies the semi-group property. The same technique can also be used for $g(t) = Q(t)s(t)$, where $Q(t)$ is a polynomial and $s(t)$ is any function satisfying the semi-group property (namely $s(t_1+t_2) = s(t_1)s(t_2)$.) By approximating a power-law function with a sum of exponentials, this technique can also be applied there. In particular, let $g(t) = \alpha_b t e^{-b t}$, where $\alpha_b$ is chosen such that $\int_0^\infty g(t)\ dt = 1$. For simplicity consider the univariate self-exciting point process with $\lambda(t), t\in (0,T],$ given by

\begin{equation}\label{cif_g_eq1}
\begin{split}
\lambda(t) &= u + a \sum_{0<t_i<t}g(t-t_i) \\
&= u + a \sum_{0<t_i<t} \alpha_b(t-t_i)e^{-b(t-t_i)}.
\end{split}
\end{equation}
Take $t\in (c_k,d_k]$ and assuming $t_i\neq c_k$, \eqref{cif_g_eq1} becomes
\begin{equation}\label{cif_g_eq2}
\begin{split}
\lambda(t) &= u + a \sum_{0<t_i<c_k} \alpha_b (t-t_i)e^{-b(t-t_i)}\\
&+ a\sum_{c_k < t_i<t} \alpha_b (t-t_i)e^{-b(t-t_i)}.
\end{split}
\end{equation}
Let $A = a \sum_{0<t_i<c_k} \alpha_b (t-t_i)e^{-b(t-t_i)}$, then
\begin{eqnarray*}
A &=&  a \sum_{0<t_i<c_k} \alpha_b (t-c_k+c_k -t_i)e^{-b(t-c_k+c_k-t_i)}\\
&=& a \sum_{0<t_i<c_k} \alpha_b (c_k-t_i)e^{-b(c_k-t_i)} \left[e^{-b(t-c_k)} \right] \\
&+& a \sum_{0<t_i<c_k} \alpha_b e^{-b(c_k-t_i)} \left[(t-c_k)e^{-b(t-c_k)} \right].
\end{eqnarray*} 
Apply the last equation to \eqref{cif_g_eq2}, we get
\begin{equation}\label{cif_g_eq3}
\begin{split}
\lambda(t) &= u + (\lambda(c_k)-u) e^{-b(t-c_k)} \\
&+ (\widetilde{\lambda}(c_k) - u)  \left[(t-c_k)e^{-b(t-c_k)} \right]\\
&+ a\sum_{c_k < t_i<t} \alpha_b (t-t_i)e^{-b(t-t_i)}, \mbox{ for } t\in (c_k,d_k],
\end{split}
\end{equation}
where $ \widetilde{\lambda}(c_k) =  u + a \sum_{0<t_i<c_k} \alpha_b e^{-b(c_k-t_i)}$ has the form of a univariate Hawkes process. In this case the extra unknowns are $\{\lambda(c_k)\}$ and $\{\widetilde{\lambda}(c_k)\}$. In general, for $g(t) = Q(t)e^{-bt}$ with $Q(t)$ being a polynomial of degree $M$, there will be $M+1$ extra unknown boundary values to solve.


\section{Numerical Results}\label{num_result}

Given the parameters $u\in \R^n,a\in R^{N\times N}$ and $b\in \R^N$, we use the algorithm from \cite{dassios2011dynamic} to simulate events $E_m\subset (0,T]$, for some $T>0$. Given the observed intervals $\{(c_{m,k},d_{m,k}]\}_{k=1}^{K_m}$ generated by Algorithm \ref{cd_alg}, the observed events are then computed as 
\begin{equation}\label{obs_eq}
O_m = E_m \cap \left(\cup_{k=1}^{K_m} (c_{m,k},d_{m,k}] \right). 
\end{equation}
In the following examples, we apply $O_m$ to MHP and MHPG given in \eqref{J_mhp} and \eqref{J_mhpg} respectively using \eqref{G_eq} for the regularization on the matrix $a$. We consider the following two constraints for the unknown boundary values $\{\lambda_{m,k}\}$:
\begin{equation}\label{cons1_eq}
\lambda_{m,k} = u_m,
\end{equation}
and
\begin{equation}\label{cons2_eq}
u_m\le \lambda_{m,k} \le 20 u_m.
\end{equation}
We use the following metrics to compare the performance of the three methods: 1) the boxplots and median values of the reconstructed parameters for $100$ simulations, and 2) the histograms of event counts on the interval $(0,20]$ having $\lambda_{m,k} = u_m$. In the latter, the median values are used to simulate events for $500$ times.

\begin{algorithm}[Algorithm for generating observed intervals]\label{cd_alg}
Given a fraction of observations $0<p<1$, and $0<\tau_1 < \tau_2$ representing the lower and upper bounds for the lengths of the observed intervals.
\begin{enumerate}
\item Set $c_{1} = 0$ and $d_1 = e_1$ where $e_1$ is a uniform random number in $(\tau_1,\tau_2)$.
\item Suppose $c_{k-1}$ and $d_{k-1}$ are computed.
\item Set $c_k = d_{k-1} + n_k$, where $n_k$ is a uniform random number in $(\frac{\tau_1}{2p},\frac{\tau_2}{2p})$.
\item Set $d_k = c_k + e_k$, where $e_k$  is a uniform random number in $(\tau_1,\tau_2)$.
\item Proceed until either $c_k \ge T$ or $d_k \ge T$.
\end{enumerate}
Set the last $d_k = T$.
\end{algorithm}

\begin{example}\label{ex1}
Consider a bivariate Hawkes process with the parameters
\begin{equation}\label{ex1_param_eq}
u = \begin{pmatrix} 5\\ 5\end{pmatrix},\ a = \begin{pmatrix}0.5 & 0.5\\ 0 & 0.5 \end{pmatrix},\ b = \begin{pmatrix}10\\ 10 \end{pmatrix}.
\end{equation}
Here, all entities have the same observed intervals which are generated by Algorithm \ref{cd_alg} with $T=1000$, $p=0.3$, $\tau_1 = 5/b_1$ and $\tau_2 = 30/b_1$. Using the observed events $O_m$ (defined in \eqref{obs_eq}), the boxplots in Figure \ref{ex1_fig1} show for 100 simulations the reconstructed parameters using MHP (top), and MHPG using \eqref{cons1_eq} (middle) and \eqref{cons2_eq} (bottom) for the boundary intensity values. MHP incorrectly estimates $u$ and $a$. See for instance the median values for $u$ and $a$ using MHP in Figure \ref{ex1_fig1}:(i),(iii). From \eqref{uvh_mean_var_eq}, $\{\lambda_2(t)\}$ with $a_{2,2} = 0.5$ and $a_{2,1}=0$ has mean $2u_2$ and variance $u_2b_2$. MHPG with the constraint \eqref{cons1_eq} imposes a small bias on boundaries $\{\lambda_{2,k}\}$ and as a result produces a slightly larger background $u_2$ in its estimation (e.g. the median value is $5.2$.) One also observes a similar effect for entity $1$ where the median value for $u_1$ is $6.1$. Other than the background parameter, both methods using MHPG produce comparable results for $a$ and $b$.

Setting the boundary value $\lambda_m(0) = u_m$, we then use the median values of the reconstructed parameters (shown in Figure \ref{ex1_fig1}) from each method to simulate events in the interval $(0,20]$. Figure \ref{ex1_fig2} shows the histograms (distributions) of event counts from each entity for $500$ simulations for MHP (green), MHPG using \eqref{cons1_eq} (cyan) and MHPG using \eqref{cons2_eq} (yellow). Compare with the ground truth (blue), MHP significantly underestimates the event counts. This is apparent since we only apply $O_m$ to MHP as oppose to $E_m$. This also shows that the knowledge of gaps (or observed intervals) is crucial in capturing the underlying distributions.
\begin{figure}
\centering
$\underset{(i)}{\includegraphics[scale=0.5]{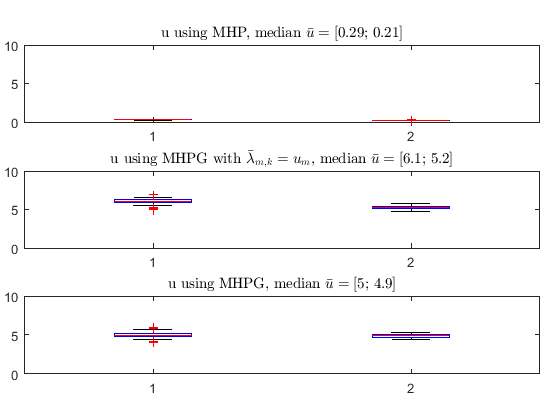}}$\\
$\underset{(ii)}{\includegraphics[scale=0.5]{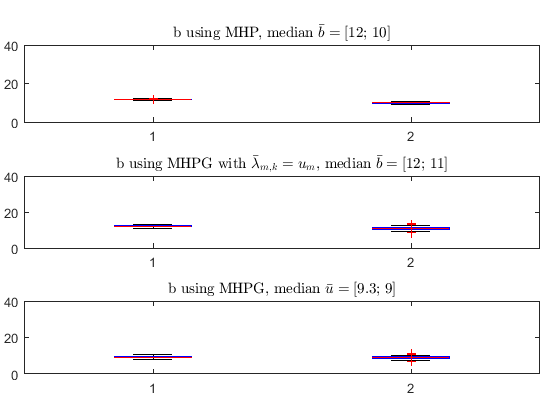}}$\\
$\underset{(iii)}{\includegraphics[scale=0.5]{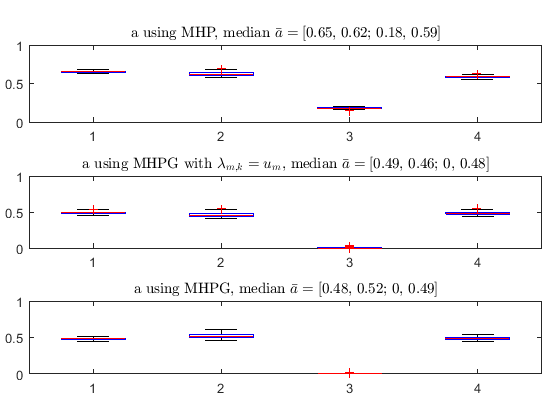}}$
\caption{Reconstructed parameters for MHP (top), MHPG using \eqref{cons1_eq} (middle) and MHPG using \eqref{cons2_eq} (bottom). The ground truths are given in \eqref{ex1_param_eq}. Here all entities have the same observed intervals generated by Algorithm \ref{cd_alg} with $T=1000$, $p=0.3$, $\tau_1 = 5/b_1$ and $\tau_2 = 30/b_1$. The median values of the $100$ reconstructions from each method are also presented in $(i), (ii)$ and $(iii)$.}
\label{ex1_fig1}
\end{figure}

\begin{figure}
\centering
$\underset{(i)}{\includegraphics[scale=0.5]{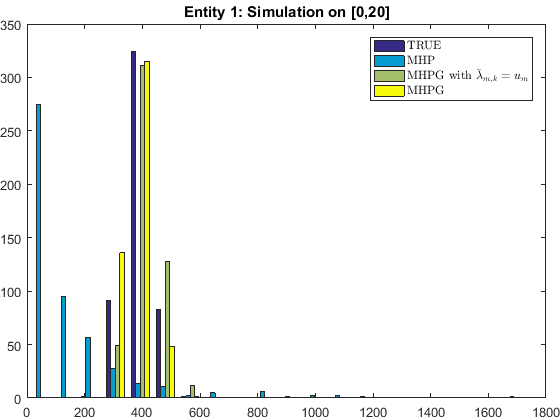}}$
$\underset{(ii)}{\includegraphics[scale=0.5]{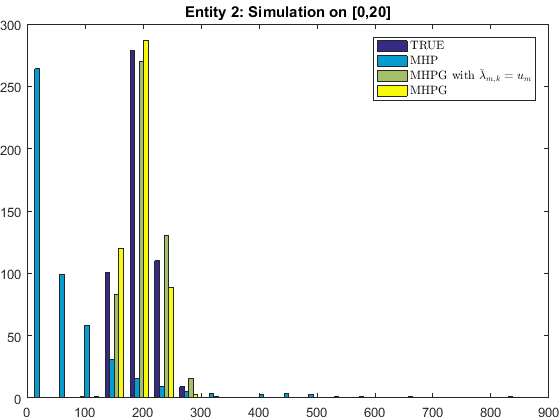}}$\\
\caption{Using the median values of the reconstructed parameters (shown in Figure \ref{ex1_fig1}) from each method to simulate events in the interval $(0,20]$ by setting the boundary value $\lambda_m(0) = u_m$, (i) and (ii) show the histograms of event counts for entity $1$ and $2$ respectively for $500$ simulations: 1) MHP (green), 2) MHPG using \eqref{cons1_eq} (cyan), and 3) MHPG using \eqref{cons2_eq} (yellow). The histograms of event counts using the true parameters are shown in blue.}
\label{ex1_fig2}
\end{figure}
\end{example}

\begin{example}\label{ex2}
The setup is the same as in Example \ref{ex1}, but here we consider a much more contagious Hawkes process having
\begin{equation}\label{ex2_gt_eq}
u = \begin{pmatrix} 1\\ 2\end{pmatrix},\ a = \begin{pmatrix}0.9 & 0.75\\ 0 & 0.9 \end{pmatrix},\ b = \begin{pmatrix}10\\ 10 \end{pmatrix}.
\end{equation}
The boxplots in Figure \ref{ex2_fig1} show, for 100 simulations, the reconstructed parameters using: MHP (top), MHPG using \eqref{cons1_eq} (middle), and MHPG using \eqref{cons2_eq} (bottom). In this case, both MHP and MHPG using \eqref{cons1_eq} (e.g. $\lambda_{m,k} = u_m$) inaccurately estimate all the parameters. However, with MHPG using \eqref{cons2_eq}, $\lambda_{m,k}$ is adjusted accordingly with respect to how the events occur in the gap $(c_{m,k},d_{m,k}]$. As a result the reconstructed parameters are significantly improved. This improvement is also validated by the histograms in Figures \ref{ex2_fig2} showing the distributions of event counts from each entity.

\begin{figure}
\centering
$\underset{(i)}{\includegraphics[scale=0.5]{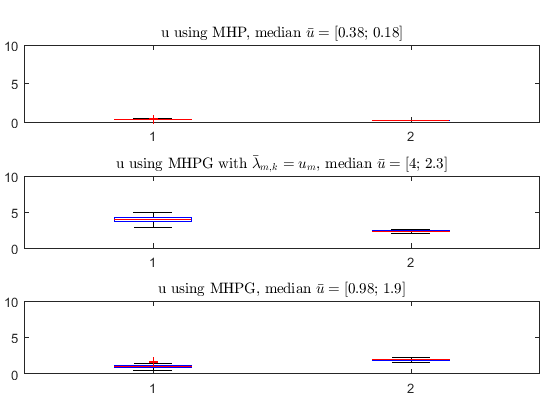}}$\\
$\underset{(ii)}{\includegraphics[scale=0.5]{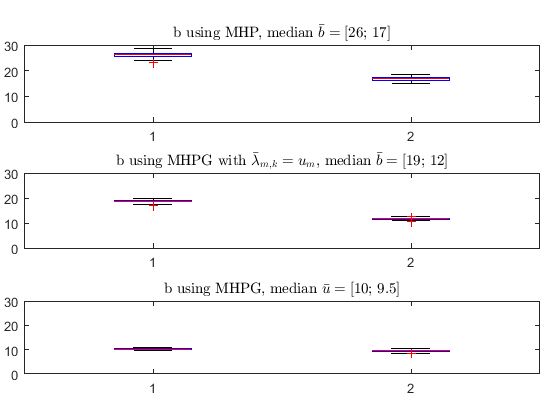}}$\\
$\underset{(iii)}{\includegraphics[scale=0.5]{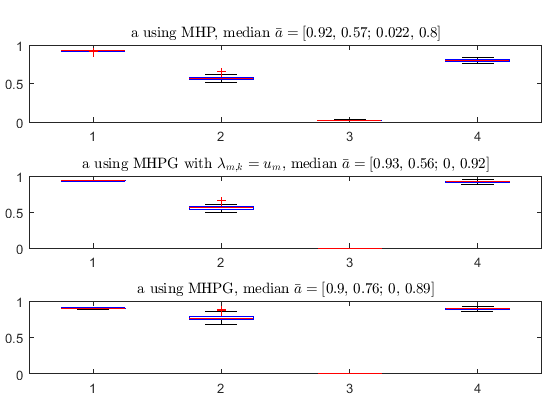}}$
\caption{Reconstructed parameters for MHP (top), MHPG using \eqref{cons1_eq} (middle) and MHPG using \eqref{cons2_eq} (bottom). The ground truths are given in \eqref{ex2_gt_eq}. Here all entities have the same observed intervals generated by Algorithm \ref{cd_alg} with $T=1000$, $p=0.3$, $\tau_1 = 5/b_1$ and $\tau_2 = 30/b_1$. The median values of the $100$ reconstructions from each method are also presented in $(i), (ii)$ and $(iii)$.}
\label{ex2_fig1}
\end{figure}

\begin{figure}
\centering
$\underset{(i)}{\includegraphics[scale=0.5]{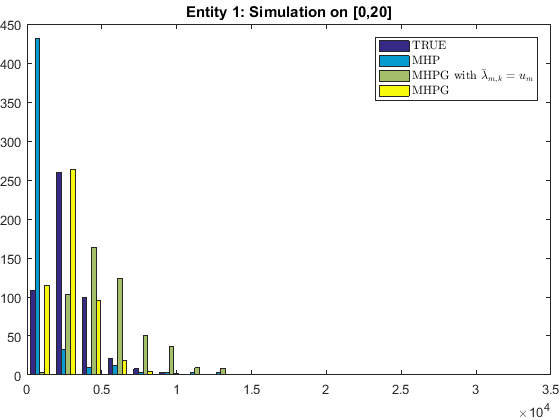}}$\\
$\underset{(ii)}{\includegraphics[scale=0.5]{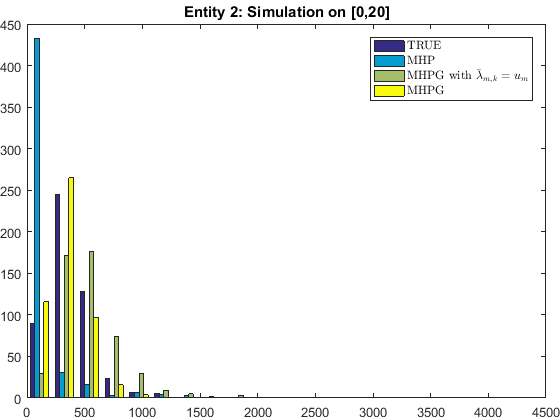}}$
\caption{Using the median values of the reconstructed parameters (shown in Figure \ref{ex2_fig1}) from each method to simulate events in the interval $(0,20]$ by setting the boundary value $\lambda_m(0) = u_m$, (i) and (ii) show the histograms of event counts for entity $1$ and $2$ respectively for $500$ simulations: 1) MHP (green), 2) MHPG using \eqref{cons1_eq} (cyan), and 3) MHPG using \eqref{cons2_eq} (yellow). The histograms of event counts using the true parameters are shown in blue.}
\label{ex2_fig2}
\end{figure}
\end{example}

\begin{example}\label{ex3}
In this example, we consider the same Hawkes process as in Example \ref{ex2}. However in this case, we generate a separate set of observed intervals for each entity using Algorithm \ref{cd_alg} with $T=1000$, $p=0.3$, $\tau_1 = 5/b_1$ and $\tau_2 = 30/b_1$. The presence of non-identical observed intervals results in all methods underestimating $a_{m,n}, m\neq n$. As seen in Figure \ref{ex3_fig1}, the estimated values for $a_{1,2}$ are much smaller than the true value of $0.75$. To compensate for this loss, all three methods overestimate $a_{1,1}$. We do not see this effect for $a_{2,2}$ since it is not influenced by entity $1$, e.g. $a_{2,1}= 0$. Among the three methods, the proposed MHPG using \eqref{cons2_eq} (bottom boxplots in Figure \ref{ex3_fig1}:(i)-(iii)) provides the closest approximations to all the parameters. This is also evident in Figures \ref{ex3_fig2} where for both entities the distributions of event counts coming from the proposed method (yellow) is closest to the ground truth (blue) compared to the other two methods (green and cyan).

\begin{figure}
\centering
$\underset{(i)}{\includegraphics[scale=0.5]{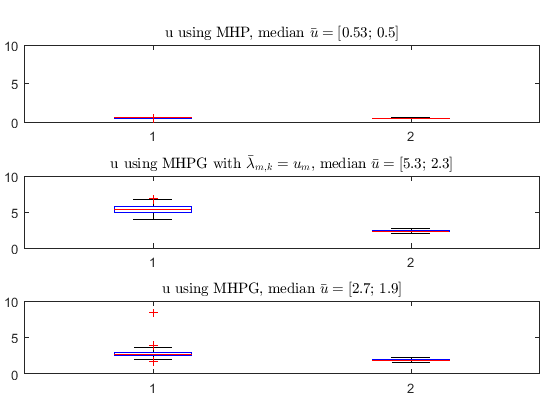}}$\\
$\underset{(ii)}{\includegraphics[scale=0.5]{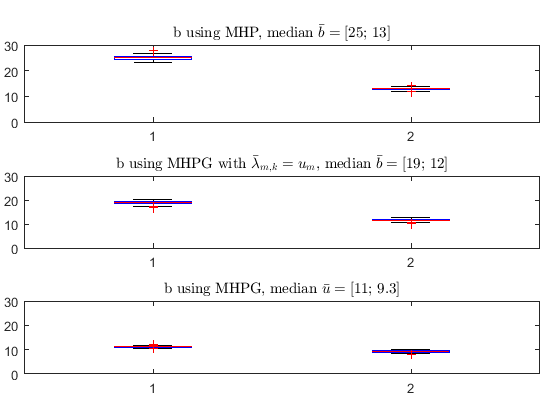}}$\\
$\underset{(iii)}{\includegraphics[scale=0.5]{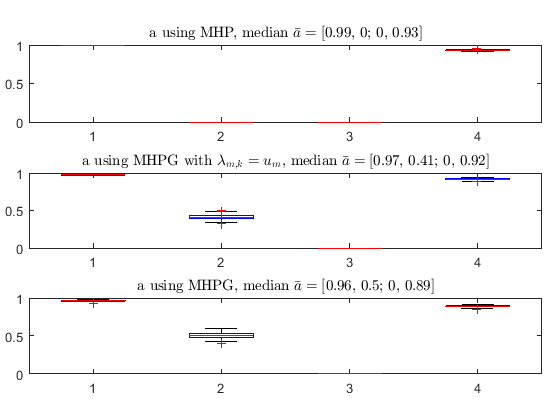}}$
\caption{Reconstructed parameters for MHP (top), MHPG using \eqref{cons1_eq} (middle) and MHPG using \eqref{cons2_eq} (bottom). The ground truths are given in \eqref{ex2_gt_eq}. Here each entity has a separate collection of observed intervals generated by Algorithm \ref{cd_alg} with $T=1000$, $p=0.3$, $\tau_1 = 5/b_1$ and $\tau_2 = 30/b_1$. The median values of the $100$ reconstructions from each method are also presented in $(i), (ii)$ and $(iii)$.}
\label{ex3_fig1}
\end{figure}

\begin{figure}
\centering
$\underset{(i)}{\includegraphics[scale=0.5]{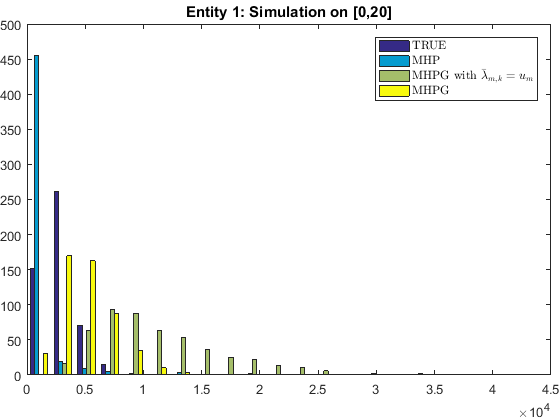}}$\\
$\underset{(ii)}{\includegraphics[scale=0.5]{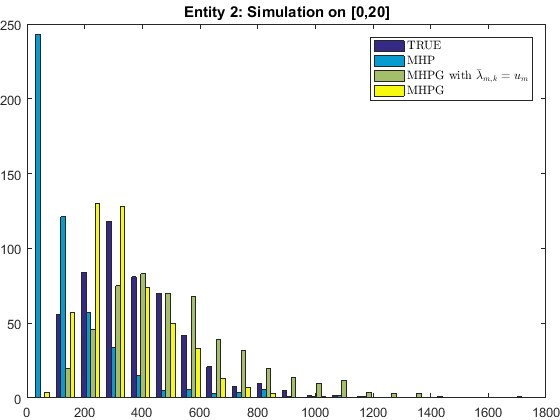}}$
\caption{Using the median values of the reconstructed parameters (shown in Figure \ref{ex3_fig1}) from each method to simulate events in the interval $(0,20]$ by setting the boundary value $\lambda_m(0) = u_m$, (i) and (ii) show the histograms of event counts for entity $1$ and $2$ respectively for $500$ simulations: 1) MHP (green), 2) MHPG using \eqref{cons1_eq} (cyan), and 3) MHPG using \eqref{cons2_eq} (yellow). The histograms of event counts using the true parameters are shown in blue.}
\label{ex3_fig2}
\end{figure}
\end{example}

\begin{example}\label{ex4}

In this example we consider the same Hawkes process as in Example \ref{ex3}. Given the nonidentical observed intervals generated as in Example \ref{ex3}, we then take the intersection of these intervals to get the common observed intervals for each entity. Figure \ref{ex4_fig0} shows the histograms of the lengths of the observed intervals prior and posterior to taking the intersection. The resulting intersection contains intervals with lengths concentrated mostly below $10/b_1$ and the fraction of observations becomes $p=0.14$. Figure \ref{ex4_fig1} shows the performance of the three methods by considering only events that are contained in the intersection. Here, we note that the reconstructed parameters using the proposed method (MHPG using \eqref{cons2_eq}) are much better than those obtained in Example \ref{ex3}. See for instance the median values of $u$ and $a$ in Figures \ref{ex4_fig1}:(i),(iii). In addition, the histogram of event counts for each entity using the proposed method (yellow) in Figure \ref{ex4_fig2} is much closer the ground truth (blue) than the histogram shown in Figure \ref{ex3_fig2} from Example \ref{ex3}. However, with the presence of too many small intervals, MHPG still under estimates $a_{1,2}$, e.g. the median value is now $0.66$ as oppose to $0.5$ in Example \ref{ex3}. Recall that the ground truth for $a_{1,2}$ is $0.75$.

\begin{figure}
\centering
\includegraphics[scale=0.5]{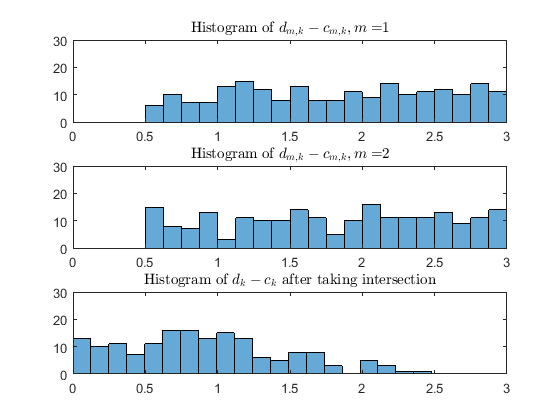}
\caption{Histograms of the lengths of the observed intervals prior and posterior to taking the intersection.}
\label{ex4_fig0}
\end{figure}

\begin{figure}
\centering
$\underset{(i)}{\includegraphics[scale=0.5]{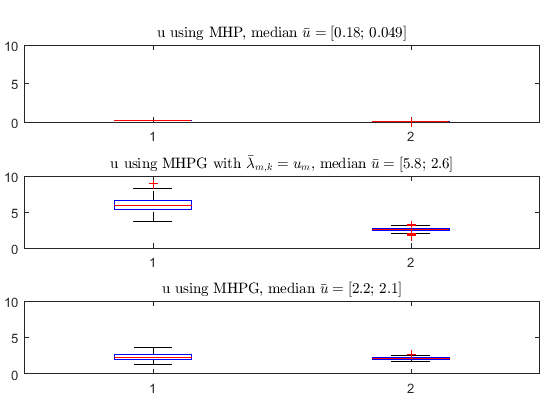}}$\\
$\underset{(ii)}{\includegraphics[scale=0.5]{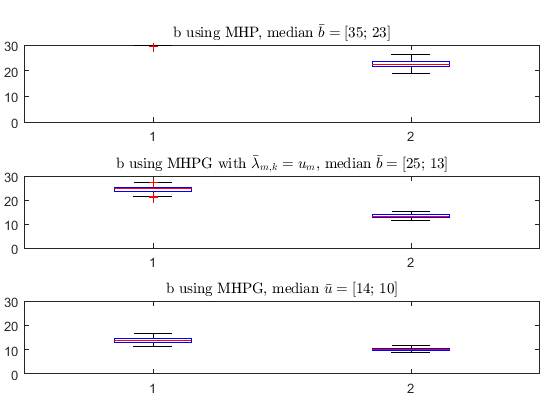}}$\\
$\underset{(iii)}{\includegraphics[scale=0.5]{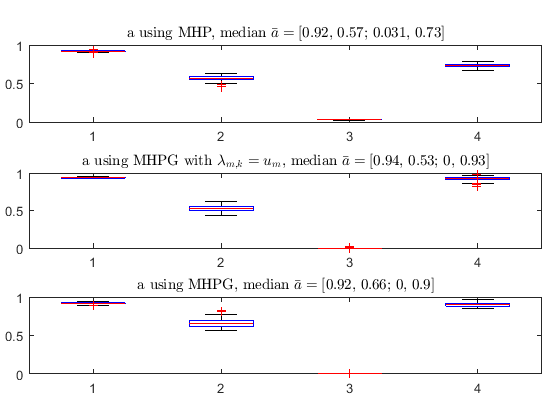}}$
\caption{Reconstructed parameters for MHP (top), MHPG using \eqref{cons1_eq} (middle) and MHPG using \eqref{cons2_eq} (bottom). The ground truths are given in \eqref{ex2_gt_eq}. Here we take intersection of observed intervals from Example \ref{ex3} and remove events that do not belong to the intersection. The median values of the $100$ reconstructions from each method are also presented in $(i), (ii)$ and $(iii)$.}
\label{ex4_fig1}
\end{figure}

\begin{figure}
\centering
$\underset{(i)}{\includegraphics[scale=0.5]{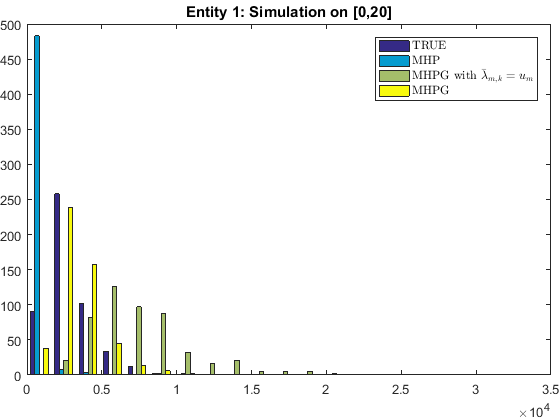}}$\\
$\underset{(ii)}{\includegraphics[scale=0.5]{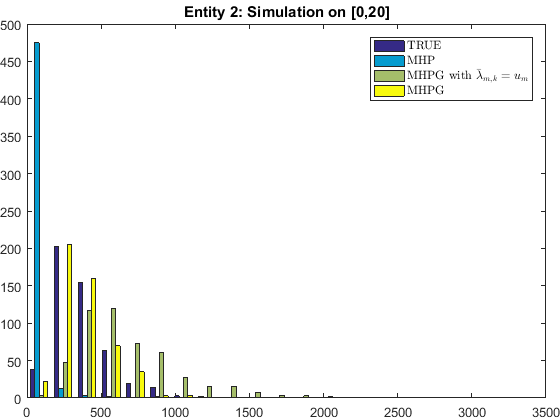}}$
\caption{Using the median values of the reconstructed parameters (shown in Figure \ref{ex4_fig1}) from each method to simulate events in the interval $(0,20]$ by setting the boundary value $\lambda_m(0) = u_m$, (i) and (ii) show the histograms of event counts for entity $1$ and $2$ respectively for $500$ simulations: 1) MHP (green), 2) MHPG using \eqref{cons1_eq} (cyan), and 3) MHPG using \eqref{cons2_eq} (yellow). The histograms of event counts using the true parameters are shown in blue.}
\label{ex4_fig2}
\end{figure}
\end{example}

\begin{example}\label{ex5}
In this example, we consider the same Hawkes process and the setup as in Example \ref{ex2} but with the fraction of observations $p=0.1$. Figures \ref{ex5_fig1}-\ref{ex5_fig2} show the reconstructed parameters and the histograms of event counts for the three compared methods. The results for MHPG using \eqref{cons2_eq} are as good as in Example \ref{ex2} where $p=0.3$. Even though the fraction of observations is smaller than the one from Example \ref{ex4}, the reconstructions are better by using uniform sampling for the observed intervals with $\tau_1 = 5/b_1$ and $\tau_2 = 30/b_1$. Again, MHPG using \eqref{cons2_eq} outperforms the other two methods.

\begin{figure}
\centering
$\underset{(i)}{\includegraphics[scale=0.5]{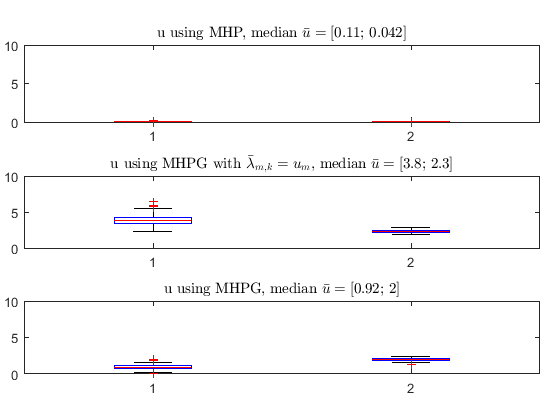}}$\\
$\underset{(ii)}{\includegraphics[scale=0.5]{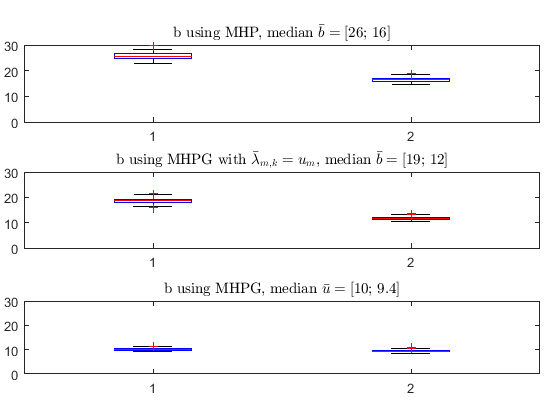}}$\\
$\underset{(iii)}{\includegraphics[scale=0.5]{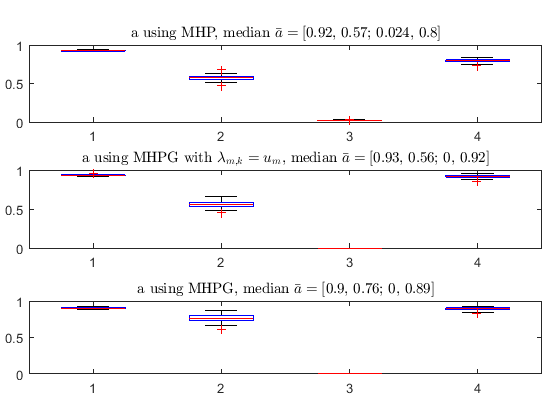}}$
\caption{Reconstructed parameters for MHP (top), MHPG using \eqref{cons1_eq} (middle) and MHPG using \eqref{cons2_eq} (bottom). The ground truths are given in \eqref{ex2_gt_eq}. Here all entities have the same observed intervals generated by Algorithm \ref{cd_alg} with $T=1000$, $p=0.1$, $\tau_1 = 5/b_1$ and $\tau_2 = 30/b_1$. The median values of the $100$ reconstructions from each method are also presented in $(i), (ii)$ and $(iii)$.}
\label{ex5_fig1}
\end{figure}

\begin{figure}
\centering
$\underset{(i)}{\includegraphics[scale=0.5]{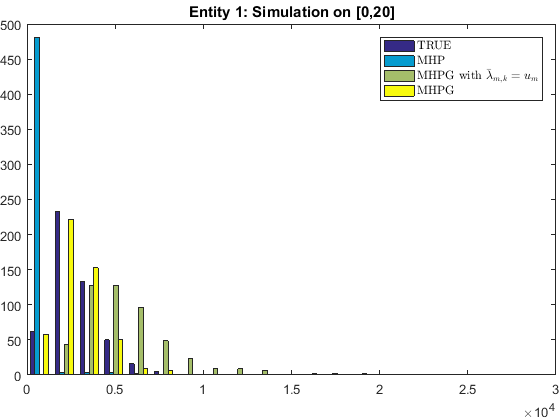}}$\\
$\underset{(ii)}{\includegraphics[scale=0.5]{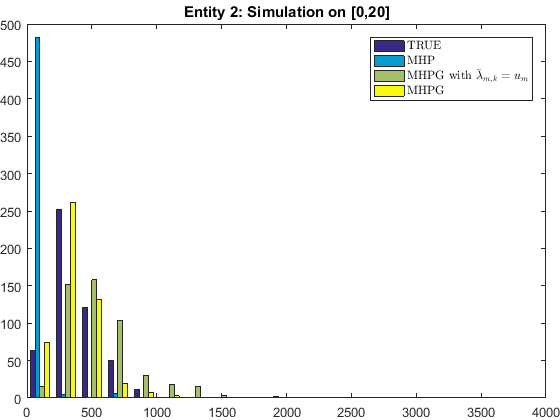}}$
\caption{Using the median values of the reconstructed parameters from each method (shown in Figure \ref{ex1_fig1}) to simulate events in the interval $(0,20]$ by setting the boundary value $\lambda_m(0) = u_m$, (i) and (ii) show the histograms of event counts from each method for entity $1$ and $2$ respectively for $500$ simulations: 1) MHP (green), 2) MHPG using \eqref{cons1_eq} (cyan), and 3) MHPG using \eqref{cons2_eq} (yellow). The histograms of event counts using the true parameters are shown in blue.}
\label{ex5_fig2}
\end{figure}
\end{example}

In conclusion, we present here a simple technique for modeling the CIF of a multivariate Hawkes process that incorporates observational gaps in \eqref{pcif_eq}. The proposed minimizing energy \eqref{J_mhpg} simplifies the problem of having to deal with a large amount of missing events by introducing a much smaller number of unknown boundary values, e.g. $\{\lambda_{m,k}\}$. In our numerical study, a constraint such as \eqref{lambda_cons} with $C\gg 1$ is sufficient for a stable reconstruction of the underlying parameters given that the observed intervals are sampled appropriately. In this paper, we consider a random uniform sampling strategy (Algorithm \ref{cd_alg}) that is based solely on the timescale $1/b_m$. However, the dynamic of $\{\lambda_m(t)\}$ is also sensitive to the parameters $u_m$ and $\{a_{m,n}\}_{n=1}^N$. Thus a sampling strategy should also take into account these additional parameters. 

The comparisons between Examples $4$ and $5$ show that a good sampling strategy is crucial as a prior step. However, in the case when the observed intervals from all entities within a network are not identical, taking the intersection (reducing the fraction of observations and observed events) can improve the parameter estimations. In a network of $N$ entities with nonidentical observed intervals, taking the intersection among all $N$ entities, where $N$ is large, may produce an empty set or a very small fraction of observations. Therefore, the techniques proposed here are not applicable to this situation. One possible heuristic approach to mitigate this problem is to consider the following strategy: Fix an $m$ and let $I_m = \{n\in \{1,\cdots,N\}: \widetilde{a}_{m,n} > 0\}$, where $\widetilde{a}_{m,n}$ is estimated by applying the intersection of only entities $m$ and $n$ to MHPG. The new CIF for $m$ can now be estimated as
\begin{equation}\label{new_cif_eq}
\begin{split}
\bar{\lambda}_m(t) &= u_m + (\lambda_m(c_k)-u_m)e^{-b_m (t-c_k)}\\
&+ \sum_{n\in I_m} a_{m,n} \sum_{c_k<t_{n,j}<t} b_m e^{-b_m(t-t_{n,j})},
\end{split}
\end{equation}  
where $\{(c_k,d_k]\}$ is the intersection between $\{(c_{m,k},d_{m,k}]\}$ and $\{(c_{n,k},d_{n,k}]\}$ for all $n\in I_n$. For a network with sparse interactions, this could provide heuristically a strategy for computing the parameters associated to entity $m$, e.g. apply the new estimated $\bar{\lambda}_m(t)$ in \eqref{new_cif_eq} to the minimizing energy in \eqref{J_mhpg}. Again, it may be possible that either the resulting $\{(c_k,d_k]\}$ does not exists (since it is an empty set) or that it is very small. So in this case, the proposed method is not applicable for this type of sampling strategy. 


\appendix

Denote $\bar{\lambda}_{m,k,i} = \bar{\lambda}_m(t_{m,i})$ for some $t_{m,i}\in (c_{m,k},d_{m,k}]$. Also, let 
$$
\bar{\Lambda}_{m,k} = \int_{c_{m,k}}^{d_{m,k}}\bar{\lambda}_m(t)\ dt.
$$
We have
\begin{eqnarray*}
\bar{\Lambda}_{m,k} &=& u_m(d_{m,k} - c_{m,k}) \\
&+& \frac{\bar{\lambda}_{m,k} - u_{m}}{b_m} \left(1 - e^{-b_m(d_{m,k}-c_{m,k})}\right)\\
&+& \sum_{n=1}^N a_{m,n} B_{m,n,k},\mbox{ and }\\
\bar{\lambda}_{m,k,i} &=& u_m + (\bar{\lambda}_{m,k} - u_{m,k})e^{-b_m(t_{m,i} - c_{m,k})} \\
&+& \sum_{n=1}^N a_{m,n}A_{m,n,k,i},
\end{eqnarray*}
where
\begin{equation}\label{amnki_eq}
\begin{split}
A_{m,n,k,i} &= \sum_{c_{m,k} < t_{n,j}<t_{m,i}\le d_{m,k}}b_m e^{-b_m(t_{m,i}-t_{n,j})}\\
&= A_{m,n,k,i-1} e^{-b_m(t_{m,i}-t_{m,i-1})}\\
&+ \sum_{c_{m,k}< t_{m,i-1}\le t_{n,j}<t_{m,i}} b_m e^{-b_m(t_{m,i}-t_{n,j})},
\end{split}
\end{equation}
which can be computed recursively and
$$
B_{m,n,k} = \sum_{c_{m,k} < t_{n,j}\le d_{m,k}} \left(1 - e^{-b_m(d_{m,k}-t_{n,j})}\right).
$$

The minimizing energy we are interested in is:
\begin{equation}\label{J_eq}
\begin{split}
&J(u,a,b,\bar{\lambda}) = \mu \sum_{m,n=1}^N |a_{m,n}| \\
&+ \sum_{m=1}^N \sum_{k=1}^{K_m}\left[ \bar{\Lambda}_{m,k}
- \sum_{t_{m,i}\in (c_{m,k},d_{m,k}]} \log(\bar{\lambda}_{m,k,i})\right]\\
&= G(a) + L(u,a,b,\bar{\lambda}),
\end{split}
\end{equation}
with the constraint that $u_m \le \bar{\lambda}_{m,k}\le Cu_m$.

{\bf Computing $u_m$:} We have
\begin{equation*}
\frac{\d J}{\d u_m} = \sum_{k=1}^{K_m}\left[ \frac{\d \bar{\Lambda}_{m,k}}{\d u_m} - \sum_{t_{m,i} \in (c_{m,k},d_{m,k}]} \frac{1}{\bar{\lambda}_{m,k,i}}\frac{\d \bar{\lambda}_{m,k,i}}{\d u_m}\right],
\end{equation*}
where
\begin{equation*}
\frac{\d \bar{\Lambda}_{m,k}}{\d u_m} = \frac{1}{b_m}\left[ e^{-b_m(d_{m,k}-c_{m,k})} - (1 - b_m(d_{m,k} - c_{m,k})\right]
\end{equation*}
and
\begin{equation*}
\frac{\d \bar{\lambda}_{m,k,i}}{\d u_m} = \left(1 - e^{-b_m(t_{m,i} - c_{m,k})}\right).
\end{equation*}
Setting $\frac{\d J}{\d u_m}=0$, we see that a minimizer $u_m$ must satisfy
\begin{equation}\label{u_eq}
u_m = \left[\sum_{k=1}^{K_m} \frac{u_m}{\bar{\lambda}_{m,k,i}}\frac{\d \bar{\lambda}_{m,k,i}}{\d u_m} \right]/\left[ \sum_{k=1}^{K_m} \frac{\d \bar{\lambda}_{m,k}}{\d u_m} \right].
\end{equation}

{\bf Computing $a_{m,n}$:} We have

\begin{eqnarray*}
\frac{\d L}{\d a_{m,n}} &=& \sum_{k=1}^{K_m} \left[\frac{\d \bar{\lambda}_{m,k}}{\d a_{m,n}} - \sum_{t_{m,i}\in (c_{m,k},d_{m,k}]} \frac{1}{\bar{\lambda}_{m,k,i}}\frac{\d\bar{\lambda}_{m,k,i}}{\d a_{m,n}}\right]\\
&=& \sum_{k=1}^{K_m}\left[ B_{m,n,k} - \sum_{t_{m,i}\in (c_{m,k},d_{m,k}]} \frac{A_{m,n,k,i}}{\bar{\lambda}_{m,k,i}}\right]\\
&=& \sum_{k=1}^{K_m} B_{m,n,k} \\
&-& \frac{1}{a_{m,n}}\sum_{k=1}^{K_m}\sum_{t_{m,i}\in (c_{m,k},d_{m,k}]} \frac{a_{m,n}A_{m,n,k,i}}{\bar{\lambda}_{m,k,i}}.
\end{eqnarray*}
Setting 
$$
\bar{a}_{m,n} = \left[\sum_{k=1}^{K_m}\sum_{t_{m,i}\in (c_{m,k},d_{m,k}]} \frac{a_{m,n}A_{m,n,k,i}}{\bar{\lambda}_{m,k,i}}\right]/\left[ \sum_{k=1}^{K_m} B_{m,n,k} \right]
$$
We then solve
\begin{equation}\label{a_eq}
a_{m,n} = shrink_{\mu}\left(\bar{a}_{m,n} \right).
\end{equation}

{\bf Computing $b_m$:} We have

\begin{eqnarray*}
\frac{\d J}{\d b_m} &=& \sum_{k=1}^{K_m}\left[\frac{\d \bar{\lambda}_k}{\d b_m} - \sum_{t_{m,i}\in (c_{m,k},d_{m,k}]}\frac{1}{\bar{\lambda}_{m,k,i}}\frac{\d\bar{\lambda}_{m,k,i}}{\d b_m} \right], 
\end{eqnarray*}
where
\begin{eqnarray*}
\frac{\d \bar{\lambda}_k}{\d b_m} &=& (\bar{\lambda}_{m,k} - u_m)\Big[-\frac{1}{b_m^2}(1 - e^{-b_m(d_{m,k}-c_{m,k})})\\
&+& \frac{1}{b_m}(d_{m,k} - c_{m,k}) e^{-b_m(d_{m,k}-c_{m,k})} \Big] \\
&+& \sum_{n=1}^N a_{m,n} \frac{\d B_{m,n,k}}{\d b_m}.
\end{eqnarray*}
and
\begin{eqnarray*}
\frac{\d \bar{\lambda}_{m,k,i}}{\d b_m} &=& - (\bar{\lambda}_{m,k}-u_m)(t_{m,i} - c_{m,k}) e^{-b_m(t_{m,i} - c_{m,k})} \\
&+& \sum_{n=1}^N a_{m,n} \frac{\d A_{m,n,k,i}}{\d b_m}
\end{eqnarray*}
where
\begin{eqnarray*}
& & \frac{\d A_{m,n,k,i}}{\d b_m} = \sum_{c_{m,k} < t_{n,j}<t_{m,i}\le d_{m,k}} e^{-b_m(t_{m,i}-t_{n,j})} \\
&-& b_m \sum_{c_{m,k} < t_{n,j}<t_{m,i}\le d_{m,k}}(t_{m,i}-t_{n,j}) e^{-b_m(t_{m,i}-t_{n,j})} \\
&=& \frac{\d A_{m,n,k,i}^{(1)}}{\d b_m} - b_m \frac{\d A_{m,n,k,i}^{(2)}}{\d b_m}.
\end{eqnarray*}
Thus,
\begin{eqnarray*}
\frac{\d J}{\d b_m} &=& \sum_{k=1}^{K_m} \frac{\d\bar{\lambda}_{m,k}}{\d b_m} \\
&-&\sum_{k=1}^{K_m}\sum_{t_{m,i}\in(c_{m,k},d_{m,k}]} \frac{1}{\bar{\lambda}_{m,k,i}}\cdot\\
& &\left(- (\bar{\lambda}_{m,k}-u_m)(t_{m,i} - c_{m,k}) e^{-b_m(t_{m,i} - c_{m,k})}\right.\\
& &\left.+ \sum_{n=1}^N a_{m,n} \frac{\d A_{m,n,k,i}^{(1)}}{\d b_m} \right)\\
&-&b_m \sum_{k=1}^{K_m} \sum_{t_{m,i}\in(c_{m,k},d_{m,k}]} \frac{\left(\sum_{n=1}^N a_{m,n} \frac{\d A_{m,n,k,i}^{(2)}}{\d b_m} \right)}{\bar{\lambda}_{m,k,i}}\\
&=& A_1 - A_2 - b_m A_3.,
\end{eqnarray*}
Setting $\frac{\d J}{\d b_m} = 0$ implies that a minimizer $b_m$ must satisfy,
\begin{equation}\label{b_eq}
b_m = \frac{A_1 - A_2}{A_3}.
\end{equation}

{\bf Computing $\bar{\lambda}_{m,k}$:}
\begin{eqnarray*}
\frac{\d J}{\d\bar{\lambda}_{m,k}} &=& \frac{1}{b_m} (1 - e^{-b_m(d_{m,k} - c_{m,k})}) \\
&-& \sum_{t_{m,i} \in (c_{m,k},d_{m,k}]} \frac{1}{\bar{\lambda}_{m,k,i}} e^{-b_m (t_{m,i} - c_{m,k})}.
\end{eqnarray*}
Setting $\frac{\d J}{\d\bar{\lambda}_{m,k}}=0$ implies that a minimizer $\bar{\lambda}_{m,k}$ must satisfy
\begin{equation}\label{lambda_eq}
\begin{split}
\bar{\lambda}_{m,k} = &b_m\left[\sum_{t_{m,i}\in(c_{m,k},d_{m,k}]} \frac{\bar{\lambda}_{m,k}e^{-b_m(t_{m,i}-c_{m,k})}}{\bar{\lambda}_{m,k,i}} \right]/\\
& \left[1 - e^{-b_m(d_{m,k}-c_{m,k})} \right],
\end{split}
\end{equation}
with the constraint $u_m\le \bar{\lambda}_{m,k} \le Cu_m$, for some $C \ge 1$.

\bibliographystyle{IEEEtran}
\bibliography{biblio}

\begin{thebibliography}{10}
\providecommand{\url}[1]{#1}
\csname url@samestyle\endcsname
\providecommand{\newblock}{\relax}
\providecommand{\bibinfo}[2]{#2}
\providecommand{\BIBentrySTDinterwordspacing}{\spaceskip=0pt\relax}
\providecommand{\BIBentryALTinterwordstretchfactor}{4}
\providecommand{\BIBentryALTinterwordspacing}{\spaceskip=\fontdimen2\font plus
\BIBentryALTinterwordstretchfactor\fontdimen3\font minus
  \fontdimen4\font\relax}
\providecommand{\BIBforeignlanguage}[2]{{%
\expandafter\ifx\csname l@#1\endcsname\relax
\typeout{** WARNING: IEEEtran.bst: No hyphenation pattern has been}%
\typeout{** loaded for the language `#1'. Using the pattern for}%
\typeout{** the default language instead.}%
\else
\language=\csname l@#1\endcsname
\fi
#2}}
\providecommand{\BIBdecl}{\relax}
\BIBdecl

\bibitem{ogata1984transition}
Y.~Ogata and K.~Shimazaki, ``Transition from aftershock to normal activity: The
  1965 rat islands earthquake aftershock sequence,'' \emph{Bulletin of the
  Seismological Society of America}, vol.~74, no.~5, pp. 1757--1765, 1984.

\bibitem{ogata1988statistical}
Y.~Ogata, ``Statistical models for earthquake occurrences and residual analysis
  for point processes,'' \emph{Journal of the American Statistical
  Association}, vol.~83, no. 401, pp. 9--27, 1988.

\bibitem{ogata1999seismicity}
------, ``Seismicity analysis through point-process modeling: A review,''
  \emph{Pure and applied geophysics}, vol. 155, no. 2-4, pp. 471--507, 1999.

\bibitem{zhuang2002stochastic}
J.~Zhuang, O.~Yosihiko, and V.-J. D., ``Stochastic declustering of space-time
  earthquake occurrences,'' \emph{Journal of the American Statistical Society},
  vol.~97, no. 458, pp. 369--380, 2002.

\bibitem{mohler2011self}
G.~O. Mohler, M.~B. Short, P.~J. Brantingham, F.~P. Schoenberg, and G.~E. Tita,
  ``Self-exciting point process modeling of crime,'' \emph{Journal of the
  American Statistical Association}, vol. 106, no. 493, 2011.

\bibitem{lewis2012self}
E.~Lewis, G.~Mohler, P.~J. Brantingham, and A.~L. Bertozzi, ``Self-exciting
  point process models of civilian deaths in iraq,'' \emph{Security Journal},
  vol.~25, no.~3, pp. 244--264, 2012.

\bibitem{sidebottom2012repeat}
A.~Sidebottom, ``Repeat burglary victimization in malawi and the influence of
  housing type and area-level affluence,'' \emph{Security journal}, vol.~25,
  no.~3, pp. 265--281, 2-12.

\bibitem{short2014gang}
M.~Short, G.~Mohler, P.~Brantingham, and G.~Tita, ``Gang rivalry dynamics via
  coupled point process networks,'' \emph{Discrete \& Continuous Dynamical
  Systems-Series B}, vol.~19, no.~5, 2014.

\bibitem{ait2010modeling}
Y.~A{\"\i}t-Sahalia, J.~Cacho-Diaz, and R.~J. Laeven, ``Modeling financial
  contagion using mutually exciting jump processes,'' National Bureau of
  Economic Research, Tech. Rep., 2010.

\bibitem{azizpour2008self}
S.~Azizpour and K.~Giesecke, ``Self-exciting corporate defaults: Contagion vs.
  frailty,'' Stanford University working paper series, Tech. Rep., 2008.

\bibitem{bowsher2007modeling}
C.~G. Bowsher, ``Modeling security market events in continuous time: intensity
  based, multivariate point process models,'' \emph{Journal of Econometrics},
  vol. 141, no.~2, pp. 876--912, 2007.

\bibitem{embrechts2011multivariate}
P.~Embrechts, T.~Liniger, L.~Lin \emph{et~al.}, ``Multivariate hawkes
  processes: an application to financial data,'' \emph{Journal of Applied
  Probability}, vol.~48, pp. 367--378, 2011.

\bibitem{embrechts2016hawkes}
P.~Embrechts and M.~Kirchner, ``Hawkes graphs,'' \emph{arXiv preprint
  arXiv:1601.01879}, 2016.

\bibitem{stomakhin2011reconstruction}
A.~Stomakhin, M.~B. Short, and A.~L. Bertozzi, ``Reconstruction of missing data
  in social networks based on temporal patterns of interactions,''
  \emph{Inverse Problems}, vol.~27, no.~11, p. 115013, 2011.

\bibitem{zipkin14point}
J.~R. Zipkin, F.~P. Schoenberg, K.~Coronges, and A.~L. Bertozzi,
  ``Point-process models of social network interactions: parameter estimation
  and missing data recovery,'' \emph{UCLA CAM Report}, no. 14-65, August, 2014.

\bibitem{hall2014tracking}
E.~C. Hall and R.~M. Willett, ``Tracking dynamic point processes on networks,''
  \emph{arXiv preprint arXiv:1409.0031}, 2014.

\bibitem{masuda2013self}
N.~Masuda, T.~Takaguchi, N.~Sato, and K.~Yano, ``Self-exciting point process
  modeling of conversation event sequences,'' in \emph{Temporal
  Networks}.\hskip 1em plus 0.5em minus 0.4em\relax Springer, 2013, pp.
  245--264.

\bibitem{etesami2016learning}
J.~Etesami, N.~Kiyavash, K.~Zhang, and K.~Singhal, ``Learning network of
  multivariate hawkes processes: A time series approach,'' \emph{arXiv preprint
  arXiv:1603.04319}, 2016.

\bibitem{kim2011granger}
S.~Kim, D.~Putrino, S.~Ghosh, and E.~N. Brown, ``A {\uppercase {g}}ranger
  causality measure for point process models of ensemble neural spiking
  activity,'' \emph{PLoS computational biology}, vol.~7, no.~3, p. e1001110,
  2011.

\bibitem{reynaud2013inference}
P.~Reynaud-Bouret, V.~Rivoirard, and C.~Tuleau-Malot, ``Inference of functional
  connectivity in neurosciences via hawkes processes,'' in \emph{Global
  Conference on Signal and Information Processing (GlobalSIP), 2013
  IEEE}.\hskip 1em plus 0.5em minus 0.4em\relax IEEE, 2013, pp. 317--320.

\bibitem{daley2003introduction}
D.~J. Daley and D.~Vere-Jones, \emph{An Introduction to the Theory of Point
  Processes, volume I: Elementary Theory and Methods of Probability and its
  Applications}.\hskip 1em plus 0.5em minus 0.4em\relax Springer, New York,,
  2003.

\bibitem{hawkes1971spectra}
A.~G. Hawkes, ``Spectra of some self-exciting and mutually exciting point
  processes,'' \emph{Biometrika}, vol.~58, no.~1, pp. 83--90, 1971.

\bibitem{tibshirani1996regression}
R.~Tibshirani, ``Regression shrinkage and selection via the lasso,'' \emph{J.
  of Royal Stat. Society, Series B (Methodological)}, pp. 267--288, 1996.

\bibitem{dassios2011dynamic}
A.~Dassios and H.~Zhao, ``A dynamic contagion process,'' \emph{Advances in
  applied probability}, pp. 814--846, 2011.

\bibitem{bolte2014proximal}
J.~Bolte, S.~Sabach, and M.~Teboulle, ``Proximal alternating linearized
  minimization for nonconvex and nonsmooth problems,'' \emph{Mathematical
  Programming}, vol. 146, no. 1-2, pp. 459--494, 2014.

\bibitem{xu2014globally}
Y.~Xu and W.~Yin, ``A globally convergent algorithm for nonconvex optimization
  based on block coordinate update,'' \emph{arXiv preprint arXiv:1410.1386},
  2014.

\end{thebibliography}

\end{document}